\documentclass[11pt]{article}

\usepackage[preprint]{acl}

\usepackage{setspace}

\usepackage{times}
\usepackage{latexsym}
\usepackage[T1]{fontenc}
\usepackage[utf8]{inputenc}
\usepackage{microtype}
\usepackage{inconsolata}
\usepackage{amssymb}
\usepackage{url}
\usepackage{pifont}
\newcommand{\cmark}{\textcolor{green!50!black}{\ding{52}}}
\newcommand{\xmark}{\textcolor{red!70!black}{\ding{56}}}
\usepackage{amsmath}
\usepackage{booktabs}
\usepackage{fontawesome5}
\usepackage{cancel}
\usepackage{array}
\usepackage{multirow}
\usepackage[owncounter]{covington} 
\usepackage{etoolbox}
\BeforeBeginEnvironment{examples}{\vspace{-0.5\baselineskip}}
\AfterEndEnvironment{examples}{\vspace{-0.5\baselineskip}}
\appto\normalsize{\setlength{\abovedisplayskip}{3pt}\setlength{\belowdisplayskip}{3pt}\setlength{\abovedisplayshortskip}{0pt}\setlength{\belowdisplayshortskip}{3pt}}
\usepackage{graphicx}
\usepackage[most]{tcolorbox}
\tcbset{
  base/.style={
    arc=0mm, bottomtitle=0.5mm, boxrule=0mm, colbacktitle=black!10!white, colframe=black!30!white, coltitle=black,
    fonttitle=\bfseries\fontsize{9}{11}\selectfont, left=2.5mm, leftrule=1mm, right=3.5mm, title={#1}, toptitle=0.75mm,
    breakable,
    listing options={style=tcblatex, breakindent=0pt, breaklines=true, basicstyle=\ttfamily\fontsize{7.5}{8.5}\selectfont}
  }
}

\newtcbinputlisting[use counter=codeboxinput]{\codeboxinput}[3][]{%
  listing file={#3}, base={Listing \thetcbcounter: #2}, listing only, #1
}

\newcommand{\bad}{\#}

\usepackage[textsize=footnotesize]{todonotes}

\title{Assertion and Belief:\\ Moore's Paradox in Language Models}

\author{Anshun Asher Zheng\textsuperscript{$\clubsuit$}, Weili Cao\textsuperscript{$\diamondsuit$}, Fangcong Yin\textsuperscript{$\heartsuit$}, David Beaver\textsuperscript{$\spadesuit$, $\clubsuit$}, Junyi Jessy Li\textsuperscript{$\clubsuit$}\\ \textsuperscript{$\clubsuit$}The University of Texas at Austin,
\textsuperscript{$\diamondsuit$}Duke University\\
\textsuperscript{$\heartsuit$}New York University,
\textsuperscript{$\spadesuit$}University of Graz
         \\ \texttt{\{asher.zheng, jessy\}@utexas.edu}, \texttt{weili.cao@duke.edu} \\ \texttt{fy666@nyu.edu}, \texttt{david.beaver@uni-graz.at}} 
\begin{document}
\maketitle

\begin{abstract}
Moore's paradox \citep{moore1942} concerns sentences such as ``It is raining, but I don't believe that it is,'' which may well be true yet cannot be sensibly asserted. The absurdity lies not in the content but in the act, since to assert is to vouch for one's own belief. This makes the paradox a test of whether language models take the speech act into account when computing meaning. We present the first test of Moore's paradox in LLMs, evaluating 12 models on a curated set of Moorean sentences in varied contexts. We find that models largely recognize the infelicity of Moorean sentences, and that the judgement is not tied to a memorized form and tracks the speech act. Using the Rational Speech Act framework (RSA; \citealt{goodmanfrank2016}), we then test whether these judgements arise from a listener reasoning about a speaker. We find the models' own listeners do not reason about their own speakers, and their reasoning traces suggest that they instead recall knowledge about the paradox and apply rules therein. The models thus appear to have learned the paradox as a fact about sentences, not as a requirement on speakers. \href{https://github.com/asherz720/Moore-s-paradox-in-llms}{\faGithub\ Github}
\end{abstract}

\section{Introduction}
\label{sec:intro}
I may say that \textit{it is raining} and that \textit{you do not believe it}. I may say that \textit{it rained yesterday} and that \textit{I did not then believe it}. I may even suppose, for the sake of argument, that \textit{it is raining} and that \textit{I do not believe it}. But I cannot say, in my own voice and of the present moment, ``\textit{It is raining, but I don't believe that it is.}'' Both conjuncts may well be true together, for there is plenty of rain I don't know about; and yet to say them together is absurd \citep{moore1942}. This is Moore's paradox, and sentences that produce it, such as $p \land \neg\mathrm{B}(p)$, are called Moorean \citep{greenwilliams2007}.

\begin{table}[t]
\centering
\footnotesize
\setlength{\tabcolsep}{3pt}
\begin{tabular}{@{}l>{\raggedright\arraybackslash}p{0.56\linewidth}c@{}}
\toprule
\textbf{Act} & \textbf{State} & \textbf{Felicitous} \\
\midrule
\textbf{Assertion} & It is raining; Robin is ignorant of the rain. She shares her view over the weather.& \textbf{\xmark{} No} \\
\textbf{Reading} & Robin is asked to read a line on a worksheet ``It is raining''; she does not believe it. & \textbf{\cmark{} Yes} \\
\bottomrule
\end{tabular}
\vspace{-0.5em}
\caption{In both situations the speaker, Robin, says ``It is raining, but I don't believe that it is.''}
\label{tab:intro}
\vspace{-30pt}
\end{table}

Classic accounts of meaning, such as truth-conditional semantics, fail to predict this absurdity as the sentence is satisfiable, and a satisfiable sentence is, for them, a sound one. What these accounts overlook, Moore's paradox reveals, is that in saying something a speaker also performs an act with meaning of its own. To assert $p$ is to represent oneself as standing in an epistemic relation to $p$, as having grounds for it or believing it \citep{williamson2000,pagin2025}. In a Moorean sentence, the speaker's own self-ascription then represents her otherwise, and the two clash \citep{moore1942,williams2023}. Reading a sentence aloud, by contrast, requires only that the speaker say what is printed, not that she believe it, so the same words cause no clash (Table~\ref{tab:intro}). The paradox thus brings out what a competent language user must grasp beyond literal meaning, i.e., what a speech act requires of the speaker (e.g., for assertion, the epistemic state she should be in).

Large language models (LLMs) have shown competence with linguistic form \citep{mahowald2024}, semantics, such as modality and aspect \citep{holliday2024,ma2026}, and pragmatics, such as conversational maxims and implicature \citep{hu2023pragmatic,ruis2023}. A separate line of work probes their epistemic understanding such as whether they distinguish between belief, knowledge, and fact, 
and whether they track the beliefs of themselves and others \citep{zhu2024,shapira2024,suzgun2025,trott2026}. Yet how epistemic understanding enters into meaning has been little studied. Moore's paradox offers this bridge, since the paradox arises at the level of the speech act, which links the meaning of language to the speaker's epistemic state.


We present the first test of Moore's paradox in LLMs, with a carefully curated set of Moorean sentences, each placed in contexts. First, across 12 open-weight and proprietary models, we treat each model as a listener and test how the model understands the Moorean sentence in three ways: (1) whether the model recognizes the infelicity out of context (Section~\ref{sec:baseline}), (2) whether the judgement relies on memorized form, by testing Moorean sentences of different forms with the same meaning (Section~\ref{sec:form-test}), and (3) whether the judgement responds to the speech act, by placing the speaker in a context where she reads the sentence rather than asserts it (Section~\ref{sec:context-test}, Table~\ref{tab:intro}). 

Second, we test whether these judgements follow from reasoning about the speaker's epistemic state. We adopt the Rational Speech Act framework (RSA; \citealp{goodmanfrank2016}), which allows us to model a speaker who chooses what to say given her epistemic state and a listener who infers that state from what she says. We construct states and utterances that are relevant to the paradox (Section \ref{sec:rsa}-\ref{sec:derivation}).  We first elicit the model's speaker by asking which utterances it could say in each state ($S_1(u|s)$). To find what norms of assertion this speaker follows, we compare the speaker with alternative ones built from different norms, e.g., asserting only what one believes (BNA) or only what one's evidence supports (JNA) \citep{pagin2025}. We then elicit the model's listener separately by asking which states a speaker could be in when she says each utterance ($L_1(s|u)$). If the model reasons about the speaker, its listener should consider only the states where its own speaker would say the utterance, and we measure how well the two match. We additionally examine recurring patterns in the models' reasoning traces (Appendix~\ref{app:reasoning-traces}).

Our results show that models do \emph{not} judge Moorean sentences by reasoning about its own speaker, even though they largely recognize the pragmatic profile of them. (1) \textbf{Models recognize the pragmatic profile of Moorean sentences.} They rank the Moorean sentence between the coherent sentence and the contradiction. The judgement is insensitive to the forms of Moorean sentences while being sensitive to the speech act change in the context (Sections~\ref{sec:baseline} and~\ref{sec:form}). (2) \textbf{Their speakers follow a belief norm of assertion.} They assert what they believe even against their evidence. When the speaker's evidence shows $p$ but she does not believe $p$, the models' speakers would only assert $\neg p$ (Section~\ref{sec:model-speaker}). (3) \textbf{Yet their listeners do not follow their own speakers.} They take the Moorean sentence to also come from such a speaker, and reasoning makes such mismatches more frequent (Section~\ref{sec:model-listener}). (4) \textbf{Instead, the models rely on a recalled rule.} Their reasoning traces judge the sentence by applying or reinterpreting a known rule of the paradox (Appendix~\ref{app:reasoning-traces}).


\section{Background}
\label{sec:background}
\citet{moore1942} noticed that a speaker \emph{cannot} sensibly say the following:


\begin{examples}
\item \bad{} I went to the pictures last Tuesday, but I don't believe that I did. \label{ex:omissive}
\end{examples}

\noindent Hearers of such a sentence ``hear contradiction, yet are puzzled to find no contradiction in what has been asserted'' \citep{williams2023}. We study this omissive form, $p \land \neg\mathrm{B}(p)$, and set aside the commissive form $p \land \mathrm{B}(\neg p)$ and other forms.

Since Moore, the absurdity has been located in the act of assertion rather than in the content asserted. The explanation rests on a pragmatic principle: ``when a man asserts anything that might be true or false, he implies that he himself, at the time of speaking, believes or knows the thing in question'' (\citealt{moore1942}; see also \citealt{black1952,wittgenstein1980a,davidson1984,pagin2025}). Hence the first conjunct of a Moorean sentence, by the pragmatic principle, represents the speaker as believing $p$ at the time of speaking, while the second has the same speaker deny that very belief, which leads to the contradiction \citep{Moore1993-MOOGMS}. It follows that the absurdity disappears when the utterance is not an assertion, as when it is recited or read aloud \citep{williams2023}: the words then represent nothing about the speaker's epistemic state. 




\subsection{Beneath the surface: Epistemic structure of the Moorean state}

\label{sec:axes}

So what may a speaker assert? This concerns two questions: What epistemic standing does assertion require of the speaker? When can the position she expresses be rationally held?

\paragraph{Norms of assertion} Epistemic norms of assertion govern when a speaker can appropriately assert a proposition \citep{williamson2000, pagin2025}. On the account employed in \citet{lackey2007}'s treatment of Moorean assertions, a speaker should assert $p$ \textit{only if} it is reasonable for her to believe $p$ and her assertion is based at least partly on that reasonableness. Hence, a Moorean assertion requires the speaker to have grounds that support $p$ and $\neg\mathrm{B}(p)$, as in (\ref{ex:ep_structure}).
\vspace{0.1em}
\begin{examples}
    \item The skywalk is safe, but I do not believe it is.
    \label{ex:ep_structure}
\end{examples}
\noindent For a speaker to appropriately assert (\ref{ex:ep_structure}), she might have inspection results and repeated safe crossings that provide the grounds for asserting safety ($p$). In the meantime, the speaker notices that she steps back and still expects the glass to give way. Recognising these persistent doubts can support her self-ascription of non-belief ($\neg\mathrm{B}(p)$).


\paragraph{Epistemic rationality} Suppose both assertions have adequate grounds: the speaker has evidence supporting safety and accurately recognises her non-belief. A further question remains: given the grounds available to her, is it reasonable for her to remain unconvinced? If she has considered the safety evidence but persistently withholds belief without countervailing grounds, her position is \textit{irrational} \citep{feldmanconee1985, foley1987theory, filevabrakel2019}. Both claims may be assertable but the assertion is not a rational one. \citet{filevabrakel2019} argue that some Moorean positions can be rationally maintained. In the skywalk case, engineering evidence supports the assertion that the bridge is safe, yet the bridge looks dangerous to her. Such an appearance is a \textit{first-person seeming}, the way things strike the speaker herself through automatic processes such as perception and learned association. The seeming sustains her non-belief. Recognising the doubt supports her self-ascription without necessarily undermining the grounds for asserting safety. 

\subsection{Related Work}
\label{sec:related}

The paradox concerns three issues: meaning beyond truth-conditions (e.g., pragmatics), epistemic understanding, and speech acts. LLMs have been tested on many pragmatic inferences, from implicature and presupposition cast as natural language inference \citep{jeretic2020} to broader benchmarks and comparisons with human judgements \citep{hu2023pragmatic,ruis2023,sravanthi2024,ma2025survey}. A separate line of work evaluates whether models track others' false beliefs \citep{ullman2023, kosinski2024,shapira2024,trott2026}, represent the beliefs of themselves and others \citep{zhu2024}, and distinguish belief from knowledge and fact \citep{suzgun2025}. 
Yet thus far, pragmatic inference tests rarely probe the speaker's epistemic state, at least not directly. The domain of speech acts is where the two lines of work can intersect. Work on LLMs and speech acts has remained philosophical, debating whether a model can assert at all \citep{gubelmann2024,williamsbayne2026} rather than whether a model takes the speech act into account when computing meaning. Moorean sentences have been tested only with human participants \citep{vanelswyk2026,willemsen2026}.

The Rational Speech Act framework (RSA) models communication as a speaker and a listener reasoning about each other \citep{goodmanfrank2016}. Prior work has used RSA to explain LLM pragmatics or to improve model performance \citep{hu2023expectations,carenini2023,qiu2025}. For Moore's paradox, RSA lets us derive the pragmatic inference from the interaction between a speaker and a listener, and we can design this interaction around the factors the paradox concerns, such as the speaker's epistemic state. 



\section{Do models recognize the infelicity of Moorean assertions?}
\label{sec:baseline}
We first test whether models, as listeners, recognize the absurdity of classic Moorean assertions made in an underspecified context.

\paragraph{Materials.}
We construct stimuli from 200 propositions, which we select manually from LLM-generated candidates to cover diverse predicates and domains (details in Appendix~\ref{app:propositions}). For each proposition $p$, we instantiate three templates: a \textbf{Moorean} sentence (``$p$, but I don't believe that''), a \textbf{coherent} control (``$p$, but that's okay''), and a logical \textbf{contradiction} (``$p$, but it isn't'').

\paragraph{Method.}
We adapt two judgement tasks from human studies: binary sense judgements \citep{willemsen2026} and seven-point acceptability ratings \citep{vanelswyk2026}. We introduce each sentence with an underspecified context: ``Imagine that Robin sincerely made the following statement.'' Models then have to answer, ``Does what Robin says make sense?'' (Yes/No) or rate the statement's acceptability from 1 (Unacceptable) to 7 (Acceptable), where acceptable means that a native speaker could use the sentence. To reduce dependence on a particular wording, we use five versions of each question in separate prompts (Appendix~\ref{app:prompt-baseline}) and average judgements for each proposition. We report averaged Yes rates and acceptability ratings across 200 propositions, with 95\% confidence intervals from 1,000 bootstrap.

\paragraph{Models.} We evaluate eleven open-weight models from three families (Qwen3.5, K2-Horizon and Gemma 4) at different sizes, and Claude Haiku 5.5 (Appendix~\ref{app:models}, Table~\ref{tab:models}). We run each model with reasoning off and on.

\paragraph{Results.}
  
\begin{figure}[t]
\centering
\includegraphics[width=\linewidth]{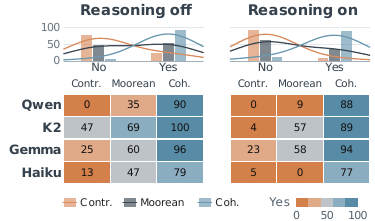}
\caption{Sensibility judgements. Top: share of No and Yes answers per sentence type, pooled over models. Bottom: share of Yes per model family. Per-model results in Appendix~\ref{app:acceptability}, Figure~\ref{fig:baseline-models}.}
\label{fig:baseline}
\vspace{-30pt}
\end{figure}

We report sense judgements (Figure~\ref{fig:baseline}) and as acceptability ratings show similar results (Appendix~\ref{app:acceptability}, Figure~\ref{fig:baseline-rating}), we report sense judgements only hereafter. As the bar charts at the top of Figure~\ref{fig:baseline} show, models overall rank the sentences coherent > Moorean > contradiction, as people do in the same sense-judgement task \citep{willemsen2026}. 
Reasoning mode lowers acceptance of the Moorean sentence in every family, but by very different amounts; see Appendix~\ref{app:reasoning-traces}, Appendix~\ref{app:acceptability}, and Figure~\ref{fig:baseline-models} for further discussions. 

\section{Underlying the Judgements}
\label{sec:form}

\begin{figure*}[t]
\centering
\includegraphics[width=\textwidth]{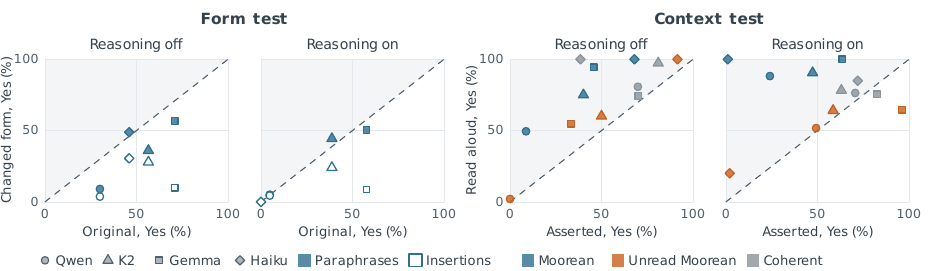}
\caption{Form test (left) and context test (right). Each point is a model family's share of Yes to ``Does what Robin says make sense?''. Left: the paraphrases and insertions against the original Moorean sentence. Right: the sentence read aloud against the sentence asserted, for the Moorean sentence, the unread Moorean control (``It's my turn, but I don't believe that'') and the coherent sentence. On the diagonal, the change makes no difference.}
\label{fig:form-context}
\end{figure*}

All models but Haiku with reasoning rank the sentences as people do (Section~\ref{sec:baseline}). We now test how they reach their judgements. A model may have memorized that sentences of the shape ``$p$, but I don't believe that'' are bad as a string. We test this in two ways: changing the form but not the situation, and the situation but not the form.

\subsection{Changing the form does not alter judgements}
\label{sec:form-test}
We keep the base prompt of Section~\ref{sec:baseline} and change only the wording of the Moorean sentence while preserving its meaning. For each Moorean sentence, we write three paraphrases that reword the second half (e.g., ``$p$, but I don't think it is'') and two insertions that place coherent material between the two halves (e.g., ``The side gate is closed, the back gate is closed, but I don't believe that the side gate is closed''; full set in Appendix~\ref{app:paraphrases}). If the judgement rested on string memorization, the judgement should change once the string changes. Figure~\ref{fig:form-context} (left) shows that the judgement survives the changes of form. For every family, the changed forms are accepted no more often than the original (per model in Appendix~\ref{app:form-models}, Figure~\ref{fig:form-models}). 

\subsection{Accommodating the infelicity}
\label{sec:context-test}
We then keep the sentence fixed and change the context to one where Robin does not make an assertion but reads it aloud, a context argued to remove the oddness (adapted from \citet{williams2023}; prompts in Appendix~\ref{app:prompt-context}):
\begin{examples}
\item A teacher asks each student to read aloud the sentence on their worksheet and add a comment of their own. Robin's worksheet says ``The side gate is closed.'' It is Robin's turn. Robin says, ``The side gate is closed, but I don't believe that.'' \label{ex:read}
\end{examples}

 We compare how often models accept the Moorean sentence when Robin asserts it and when she reads it aloud. Two controls separate the speech act from the reading frame. The coherent sentence tests whether the frame raises acceptance of any sentence. The unread Moorean sentence ``It's my turn, but I don't believe that'' tests whether the frame rescues any Moorean sentence or only the one Robin reads (i.e., speech act sensitivity). In Figure~\ref{fig:form-context} (right), a point on the diagonal means the frame makes no difference, and the further a point lies above the diagonal, the more reading aloud raises acceptance (i.e., the more the model accommodates the sentence). Reading aloud raises acceptance of the Moorean sentence by 53 points on average with reasoning and 40 without, against 6 and 15 for the coherent sentence and $-4$ and 10 for the unread Moorean sentence. For most models, accommodation thus responds to whether Robin reads or asserts, not to the frame (Appendix~\ref{app:context-models}, Figure~\ref{fig:context-models}).
 

These results show that models do not judge by form alone. They accommodate the Moorean sentence when the context allows it. However, this could come from knowledge of the paradox (Appendix~\ref{app:moore} shows that e.g., Qwen family and Haiku name ``Moore'' in more than 89\% of their reasoning traces) rather than reasoning about Robin's state.
In the next section, we test how far the judgements follow from reasoning about the speaker based on rational speech act theory.


\section{Rational Speech Act Theory}
\label{sec:rsa}

The Rational Speech Act framework (RSA; \citealt{goodmanfrank2016}) models how a listener interprets an utterance by reasoning about the speaker with three components. A \emph{literal listener} ($L_0$) interprets an utterance $u$ by its literal meaning alone, checking whether $u$ is true in each state $s$:

\footnotesize
\begin{equation} L_0(s \mid u) \propto P(s)\,[\![u]\!](s), \label{eq:l0} \end{equation}

\normalsize
\noindent where $[\![u]\!](s)$ is 1 if $u$ is true in $s$ and 0 otherwise. A \emph{pragmatic speaker} ($S_1$) in state $s$ chooses among alternative utterances, preferring those that lead the literal listener to her actual state, 

\footnotesize
\begin{equation} S_1(u \mid s) \propto \exp\big(\log L_0(s \mid u)\big). \label{eq:s1} \end{equation}

\normalsize
\noindent A \emph{pragmatic listener} ($L_1$) interprets $u$ by asking which states lead such a speaker to say it:

\footnotesize
\begin{equation} L_1(s \mid u) \propto P(s)\,S_1(u \mid s). \label{eq:l1} \end{equation}
\normalsize

\begin{table}[t]
\centering
\footnotesize
\setlength{\tabcolsep}{5pt}
\begin{tabular}{llllll}
\toprule
 & World & Evidence & Belief & Act & $M$ \\
\midrule
$s_1$ & $p$ & $p$ & $p$ & asserts & \\
$s_2$ & $p$ & none & none & asserts & \\
$s_3$ & $p$ & $p$ & $\neg p$ & asserts & \checkmark \\
$s_4$ & $\neg p$ & $\neg p$ & $\neg p$ & reads ``$p$'' & \checkmark \\
\bottomrule
\end{tabular}

\caption{The four theory-driven states (\checkmark: $M$ is felicitous). Full descriptions are in Appendix~\ref{app:states}, Table~\ref{tab:model-states}.}
\label{tab:states}
\vspace{-30pt}
\end{table}

\paragraph{Interpreting Moorean Assertions via RSA}
We first show how RSA accounts for the pragmatics of Moorean sentences, using theory-driven states. 

\noindent\emph{States.} We use four states of the speaker, Robin, $s_1$-$s_4$ (Table~\ref{tab:states}). In $s_1$, her evidence, belief and the world agree. $s_2$ is Moore's classic case: Robin has no evidence and lacks belief in $p$. In $s_3$, her evidence shows $p$, but she maintains her belief $\neg p$ due to irrationality or a first-person seeming \citep{filevabrakel2019}. In $s_4$, she reads the line ``$p$'' aloud and so does not commit herself to $p$.

\noindent\emph{Utterances.} We consider four relevant utterances about $p$: the base form ($p$), the negation ($\neg p$), a statement of ignorance ($\mathit{DK}$, ``I don't know whether $p$'') and the Moorean form ($M$). The first three assert Robin's view of $p$; $M$ asserts $p$ and denies believing it.

\begin{figure}[!tb]
  \centering
  \includegraphics[width=\linewidth]{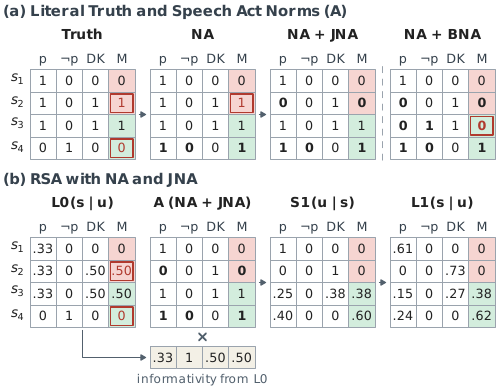}
  \caption{RSA over $s_1$-$s_4$ (uniform prior). (a) Literal truth and the norms of speech acts ($A$). The norm of reading, a non-assertive (NA) act, and then a norm of assertion (JNA, or the alternative BNA) override cells of Truth (bold). (b) RSA with NA and JNA. $S_1$ multiplies each row of $A$ by the informativity of each utterance from $L_0$ and normalizes over utterances. In the $M$ column, green marks the states where $M$ should be possible and red the states where $M$ should be ruled out (Table~\ref{tab:states}). Outlined cells mark wrong predictions.}
  
  \label{fig:rsa}
  \vspace{-30pt}
  \end{figure}

As Table~\ref{tab:states} shows, on hearing $M$, an ideal listener should consider $s_3$ and $s_4$ and rule out $s_1$ and $s_2$. We follow the computation in Figure~\ref{fig:rsa}b and derive the target profile ($L_1$ Figure~\ref{fig:rsa}b).

\emph{Literal listener.} For each state and utterance, we check against Table~\ref{tab:states} whether the utterance is true, which gives $[\![u]\!](s)$ (Truth in Figure~\ref{fig:rsa}a). $M$ is true in $s_2$ and $s_3$, where $p$ holds and Robin does not believe $p$, and false in $s_4$, where $p$ is false. The literal listener normalizes each column of Truth over states (Eq.~\ref{eq:l0}; $L_0$ in Figure~\ref{fig:rsa}b), so hearing $M$, $L_0$ splits its probability between $s_2$ and $s_3$. Literal meaning thus misses the target in $s_2$ and $s_4$.

\emph{Speaker.} A vanilla speaker may say only what is true in her state, and prefers the utterances that lead $L_0$ to it (Eq.~\ref{eq:s1}; Appendix~\ref{app:vanilla}). What she may say is thus the Truth table (Figure~\ref{fig:rsa}a). Hence, she says $M$ in $s_2$, where $M$ is true, and never in $s_4$, where $M$ is false, i.e., she cannot correct what literal meaning misses (red outlines). To reach the target, the speaker must go beyond truth: (1) the speaker may say something that is false, e.g., in $s_4$, she says $p$ although she knows it is not true (i.e., reading does not require truth); (2) the speaker may not say something even if it is true, e.g., $M$ in $s_2$, Moore's classic case.

We therefore add to the canonical $S_1$ the norms of speech acts. The speaker of Eq.~\ref{eq:s1} maximizes $\log L_0(s \mid u) = \log [\![u]\!](s) - \log \sum_{s'} [\![u]\!](s')$ (Eq.~\ref{eq:l0} with a uniform prior; derivation in Appendix~\ref{app:prior}). The first term is whether $u$ is true in her state, and the second how informative $u$ is to $L_0$. As RSA models of non-literal language revise the speaker's utility \citep{kao2014nonliteral}, we revise the first term from $\log [\![u]\!](s)$ to $\log A(s,u)$, so that what she may say goes beyond literal meaning and follows the norms of speech acts (see Appendix \ref{app:false-utt}),

\footnotesize
\begin{equation} S_1(u \mid s) \propto \exp\Big(\log A(s,u) - \log \sum_{s'} [\![u]\!](s')\Big). \label{eq:s1a} \end{equation}
\normalsize

\noindent We define $A(s,u)$ as 1 where the norm of her act lets her say $u$ in $s$ and 0 otherwise. The intuition is that the speaker maximizes informativity only among the utterances her norm permits. Where no norm applies, $A(s,u)$ is the literal $[\![u]\!](s)$, so Eq.~\ref{eq:s1a} reduces to Eq.~\ref{eq:s1}. $A$ thus depends on which act and which norm one considers, and each norm overrides some cells of the Truth table (bold in Figure~\ref{fig:rsa}a). In the case of Moore's paradox, we consider two instantiations of $A$, one for non-assertive acts and one for assertion, each with its own norm. For non-assertive acts, we instantiate the norm with reading, which requires saying what is printed and places no requirement on the speaker's epistemic state (discussion in Appendix~\ref{app:acts}). The norm of assertion instead sets what her epistemic state must be. It takes different forms \citep{williamson2000,lackey2007,pagin2025}, and we consider two of them, echoing the two submaxims of Grice's Quality \citep{grice1975}. The justification norm of assertion (JNA) says ``Assert only that for which you have proper justification'' \citep[see also][]{lackey2007}, and the belief norm of assertion (BNA) ``Assert only what you believe''. Neither instantiation of $A$ changes the literal meaning of an utterance, so $L_0$ stays the same.

With the norm of reading (NA in Figure~\ref{fig:rsa}a), the speaker in $s_4$ may say $p$, as the norm of reading requires, so $A_{\mathrm{NA}}(s_4,M) = 1$ and $S_1(M \mid s_4)$ is possible now. The reading norm does not affect her choice in $s_2$, where Robin asserts. Her choice there still rests on truth alone, so she says $M$ like the vanilla speaker, although $M$ should be ruled out (red outline). The norm of assertion corrects this. Both JNA and BNA set $A_{\mathrm{JNA}}(s_2,M) = A_{\mathrm{BNA}}(s_2,M) = 0$, JNA because Robin lacks evidence for $p$ and BNA because she lacks belief in $p$. The two norms part ways in $s_3$, where $M$ should remain possible. There her evidence shows $p$, but she believes $\neg p$. JNA lets her assert $p$, since her evidence justifies it, so $A_{\mathrm{JNA}}(s_3,M) = 1$ (NA + JNA). BNA does not, since she does not believe $p$, so $A_{\mathrm{BNA}}(s_3,M) = 0$, and $M$ is ruled out where it should be possible (NA + BNA). Figure~\ref{fig:rsa}b computes the speaker with NA and JNA. By Eq.~\ref{eq:s1a}, she weighs each utterance she may say by its informativity $1/\sum_{s'}[\![u]\!](s')$, which is $L_0(s \mid u)$ in any state where $u$ is true (.33 for $p$, 1 for $\neg p$, .50 for DK and $M$), and normalizes over utterances. In $s_4$, for instance, she may say $p$ or $M$, and so says $p$ (.40) or $M$ (.60).

\emph{Pragmatic listener.} A pragmatic listener therefore can interpret the Moorean sentence correctly by reasoning about a speaker who follows the norm of reading and JNA. Such a speaker says $M$ only in $s_3$ and $s_4$ (Eq.~\ref{eq:s1a}), so, hearing $M$, the listener considers only these two states, as the target requires ($L_1$ in Figure~\ref{fig:rsa}b).

\section{Deriving the Moorean Judgement in Models}
\label{sec:derivation}
\begin{figure*}[t]
\centering
\includegraphics[width=\textwidth]{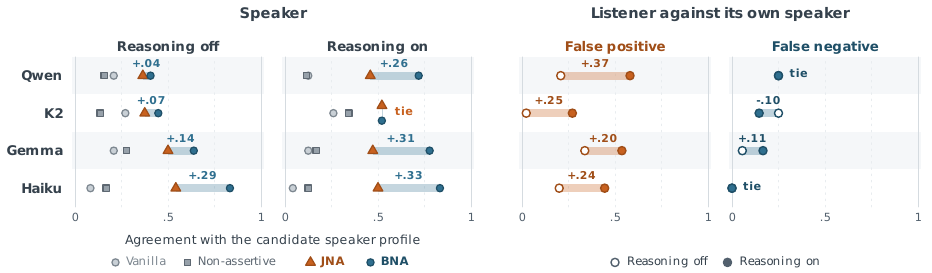}
\caption{Left: the elicited speaker's agreement with each candidate speaker; the models' speakers agree more with BNA than with JNA. Right: the false positive rate measures how often the speaker would not say $M$ while the listener considers $M$ possible, and the false negative rate measures how often the speaker would say $M$ while the listener rejects $M$. Per-model results in Appendix~\ref{app:model-grids}.}
\label{fig:speaker-listener}
\end{figure*}

Section~\ref{sec:rsa} derived the pragmatic profile of Moorean sentences with RSA. We now prompt each of the 12 LLMs separately to obtain its $\hat{L}_0$, $\hat{S}_1$ and $\hat{L}_1$, and test whether they relate as in RSA. We ask which norms of speech acts the model's $\hat{S}_1$ follows (Section~\ref{sec:model-speaker}) and whether its $\hat{L}_1$ follows from its $\hat{S}_1$ (Section~\ref{sec:model-listener}). 


\paragraph{Materials.}
\emph{Utterances.} We keep the four utterances of Section~\ref{sec:rsa}, which cover how Robin's view can relate to $p$: $p$, $\neg p$, ignorance, and the Moorean sentence. Each is instantiated with a set $Q$ of 50 of the 200 propositions of Section~\ref{sec:baseline}.

\emph{States.} Rather than fixing the states in advance, we sample them from the models. For each of the 4 utterances, we asked each of the 12 test models why Robin might have said it (prompt in Appendix~\ref{app:state-prompt}), repeating this with 3 different propositions, and collected about 10.5k proposals per utterance. We apply LLM judge for clustering each utterance's proposals, and combine with manual inspection, we arrive at 12 states in Table \ref{tab:model-states}. Details can be found in Appendix~\ref{app:proposal_cluster}.


\paragraph{Eliciting $\hat{L}_0$, $\hat{S}_1$ and $\hat{L}_1$.}
For each of the test models, we obtain their $\hat{L}_0$, $\hat{S}_1$ and $\hat{L}_1$ by prompting them as below (prompts in Appendix~\ref{app:prompt-grids}): \\
\emph{$\hat{L}_0(s|u)$}: for each state and utterance, we ask the model whether $u$ is true in $s$ ($[\![u]\!](s)$). \\

\noindent\emph{$\hat{S}_1(u|s)$}: for each state, the model rates how likely Robin is to say each of the four alternative utterances ($u'$) on the scale of 0-10. Writing $r_q(u,s)$ for the model's rating of utterance $u$ in state $s$ for proposition $q$, we normalize over utterances and average over $Q$, $\hat{S}_1(u \mid s) = \frac{1}{|Q|}\sum_{q} r_q(u,s) / \sum_{u'} r_q(u',s)$.

\noindent\emph{$\hat{L}_1(s|u)$}: for each utterance, the model rates how likely Robin was in each state when she said it on the scale of 0-10. We normalize these ratings $r_q(s,u)$ in the same way, over the 12 states, $\hat{L}_1(s \mid u) = \frac{1}{|Q|}\sum_{q} r_q(s,u) / \sum_{s'} r_q(s',u)$. Appendix~\ref{app:model-grids} reports the complete state-utterance tables for each model's elicited $\hat{L}_0$, $\hat{S}_1$, and $\hat{L}_1$.

\subsection{What kind of speaker do the models adopt?}
\label{sec:model-speaker}
We compare $\hat{S}_1$ with four candidate speakers. They differ only in which utterances Robin may say in each state ($S_1^{A}$ is Eq.~\ref{eq:s1a} with $A \in \{A_{\mathrm{Van}}, A_{\mathrm{NA}}, A_{\mathrm{JNA}}, A_{\mathrm{BNA}}\}$). (1) \emph{Vanilla} ($S_1^{\mathrm{Van}}$): Robin may say anything true. (2) \emph{Non-assertive} ($S_1^{\mathrm{NA}}$): where Robin does not assert, as when she reads aloud, teases, or misspeaks, the norm of her act rather than truth decides what she may say, so $M$ becomes possible there. (3) \emph{JNA} ($S_1^{\mathrm{JNA}}$): as (2), and where Robin asserts, she may assert only what her evidence justifies. (4) \emph{BNA} ($S_1^{\mathrm{BNA}}$): as (2), and where Robin asserts, she may assert only what she believes.


\paragraph{Method.} For each candidate speaker, we manually check, for each of the 12 states ($\mathcal{S}$) and the four utterances ($\mathcal{U}$), whether Robin may say the utterance under the candidate's norms, which gives its $A$. We then obtain $S_1^A$ by applying Eq.~\ref{eq:s1a} with $A$ to the model's own $\hat{L}_0$ (Appendix~\ref{app:model-grids}). Like the elicited $\hat{S}_1$, each candidate $S_1^A$ then decides which utterances Robin may say in each state. We compare $\hat{S}_1$, after calibration (Appendix~\ref{app:calibration}), with $S_1^A$ in each state $s$ on whether they consider $u$ feasible, writing $S_1(u \mid s) = 1$ if feasible and $0$ if not. We report the agreement as

\scriptsize
\begin{equation} \mathrm{Agrm}(\hat{S}_1, S_1^A) = \frac{1}{|\mathcal{S}||\mathcal{U}|} \sum_{s \in \mathcal{S}} \sum_{u \in \mathcal{U}} \mathbf{1}\big[\hat{S}_1(u \mid s) = S_1^A(u \mid s)\big]. \label{eq:agree} \end{equation}

\normalsize



\paragraph{Results.} \emph{The models' speakers go beyond literal truth.} No elicited speaker is closest to the vanilla speaker, except K2 7B without reasoning. For example, in $s_2$, Moore's classic case, Robin has no evidence and lacks belief in $p$. $p$ and $M$ are literally true, so the vanilla speaker would say them, yet no speaker, with or without reasoning, treats either as feasible.

\emph{Models likely adopt a BNA speaker.} Qwen, Gemma and Haiku agree most with BNA, with or without reasoning, and reasoning widens BNA's lead over JNA (Figure~\ref{fig:speaker-listener}, left). Take $s_3$, a case like the skywalk in (\ref{ex:ep_structure}). Robin has the engineering evidence that the skywalk is safe, but she does not believe it. Only BNA lets her assert that the skywalk is not safe, on belief alone and against her evidence. The models' speakers do the same. With or without reasoning, all speakers treat this assertion as feasible, except K2 3.7B without reasoning (undetermined). K2 agrees with JNA as much as with BNA with reasoning. Three of its four speakers then treat $M$ as feasible in $s_3$, which only JNA allows, while every Qwen, Gemma and Haiku speaker but Qwen 4B rules it out.

\subsection{Does the listener follow its own speaker?}
\label{sec:model-listener}
%

In RSA, the listener interprets $M$ through the speaker, $L_1(s \mid M) \propto P(s)\,S_1(M \mid s)$, so a listener that follows its own speaker considers exactly the states where its speaker would say $M$. We compare $\hat{S}_1(M \mid s)$ with $\hat{L}_1(s \mid M)$ in each of the 12 states. The listener can depart from its speaker in two ways:
(1) It can accept too much, considering a state where its speaker would never say $M$ (a \emph{false positive}); (2) It can also reject too much, ruling out $M$ in a state where its speaker would say $M$ (a \emph{false negative}; details in Appendix~\ref{app:fn}). Over the 12 states, the two rates are

\footnotesize
\[ \mathrm{FP} = \frac{\big|\{s : \hat{S}_1(M \mid s) = 0,\ \hat{L}_1(s \mid M) = 1\}\big|}{\big|\{s : \hat{S}_1(M \mid s) = 0\}\big|}, \]
\[ \mathrm{FN} = \frac{\big|\{s : \hat{S}_1(M \mid s) = 1,\ \hat{L}_1(s \mid M) = 0\}\big|}{\big|\{s : \hat{S}_1(M \mid s) = 1\}\big|}, \]
\normalsize

\noindent with $\hat{L}_1(s \mid M) = 1$ if $s$ is feasible given $M$ and $0$ if not, from the same calibration.
 We report both rates in Figure~\ref{fig:speaker-listener} (right), with each state's $\hat{S}_1$ and $\hat{L}_1$ averaged over $Q$ (per-model results in Appendix~\ref{app:model-grids}, Figures~\ref{fig:listener-unfollowed} and~\ref{fig:listener-fn}).

\paragraph{Results.} The models' listeners do not follow their own speakers, as the non-zero false positives and false negatives in Figure~\ref{fig:speaker-listener} (right) show. Only Qwen 4B, K2 7B and K2 36B-A4B without reasoning, and K2 36B-A4B with reasoning, follow their speakers in every state. The failures are mostly false positives, and reasoning boosts them. Without reasoning, the two rates are comparable (false positives .02-.34, false negatives .00-.25 per family). With reasoning, false positives rise in every family (by .20-.37) to .27-.58, while false negatives change by at most .11.

False positives fall mostly in $s_3$, $s_7$ and $s_{10}$, where Robin believes $\neg p$ although $p$ is true. In $s_3$ and $s_7$, her evidence shows $p$. The Qwen 9B-35B, Gemma and Haiku speakers rule out $M$ there and say $\neg p$ instead, as a BNA speaker would, yet their listeners take $M$ to come from $s_3$ and $s_7$, as if the speaker followed JNA. The few false negatives come from contexts such as a misspeaking Robin ($s_{12}$). Exhausted, she meant to say another sentence, and the K2 32B and Gemma 26B-A4B speakers would say $M$ there as a slip, but the models reject $M$. These results reveal a mismatch between the models' speakers and listeners.

A complementary analysis of reasoning traces suggests that models often invoke rules about assertion or reinterpret the utterance to reach their judgements (see Appendix~\ref{app:reasoning-traces}).


\section{Conclusions}
Using Moore's paradox as a test of whether language models take the speech act into account when computing meaning, we find that models largely recognize the infelicity of Moorean sentences, but their judgements do not arise from a listener reasoning about the speaker. Their speakers follow a norm of assertion, yet their listeners do not use it, and their traces point to a recalled rule instead. This reveals a gap between judging an utterance correctly and deriving the judgement from what the act requires of the speaker. The gap matters as LLMs increasingly interpret what people say, where meaning depends on who says it and in what act. We hope that deriving a model's judgements from its own speaker and listener provides a testbed for future work, including other speech acts such as promises and questions that depend on the speaker's state, and training or prompting that leads models to reason from the speaker.

\section{Limitations}
Several choices define the scope of our findings. We use templated sentences of the omissive form in English, which gives us full control over the stimuli and allows matched comparisons across the Moorean, coherent and contradictory sentences; commissive and knowledge forms, naturally occurring Moorean sentences and other languages remain for future work. We elicit each model's speaker and listener through ratings, normalized and calibrated per model. Token probabilities could complement the ratings for models without reasoning, but with reasoning the answer follows a sampled trace, so its probability reflects that one trace rather than the model's judgement, and such probabilities are not available for every model. We test 12 models across four families and sizes, with and without reasoning. Larger proprietary models may differ, although the gap we find holds across families and sizes. Finally, reasoning traces need not be faithful to the computation behind a judgement \citep{turpin2023}, so we treat them as behavior that complements the RSA analysis rather than as evidence of mechanism on their own.

\section*{Acknowledgments}

We would like to thank Kanishka Misra for the insightful discussion. This work was partially supported by NSF grants 2107524 and 2145479, as well as Good Systems,\footnote{https://goodsystems.utexas.edu/} a UT Austin Grand Challenge to develop responsible AI technologies.

\bibliography{custom}

\appendix

\section{Proposition Sampling and Selection}
\label{app:propositions}
We sample candidate scenarios using Claude Fable 5.1 (prompt in Appendix~\ref{app:prompt-propositions}). Each scenario has Robin check three objects of one kind for one property, such as three gates of an office complex and whether they are closed. Each object gives a proposition, e.g., ``The side gate is closed''. We manually select 100 scenarios to cover diverse predicates and domains, with 98 domains and 61 distinct predicates in the final set. All propositions are stative and in the simple present, describing a state of an object at the time of speaking, so that Robin can have evidence and a belief about that state. Each scenario is instantiated with two propositions, leading to a set of 200 propositions. The other two objects of the scenario supply the coherent material for the insertions of the form test (Section~\ref{sec:form}).

\section{Models}
\label{app:models}
Table~\ref{tab:models} lists the twelve test models with their sources and reasoning settings.
\begin{table*}[h]
\centering
\small
\setlength{\tabcolsep}{3pt}
\begin{tabular}{@{}llll@{}}
\toprule
Model & Parameters & Source & Reasoning on \\
\midrule
Qwen3.5-4B & 4B & \texttt{Qwen/Qwen3.5-4B} & native thinking \\
Qwen3.5-9B & 9B & \texttt{Qwen/Qwen3.5-9B} & native thinking \\
Qwen3.5-27B & 27B & \texttt{Qwen/Qwen3.5-27B} & native thinking \\
Qwen3.5-35B-A3B & 35B (3B active) & \texttt{Qwen/Qwen3.5-35B-A3B} & native thinking \\
\midrule
K2-Horizon-3.7B & 3.7B & \texttt{IFM/K2-Horizon-3.7B} & effort high \\
K2-Horizon-7B & 7B & \texttt{IFM/K2-Horizon-7B} & effort high \\
K2-Horizon-32B & 32B & \texttt{IFM/K2-Horizon-32B} & effort high \\
K2-Horizon-MoVA-36B-A4B & 36B (4B active) & \texttt{IFM/K2-Horizon-MoVA-36B-A4B} & effort high \\
\midrule
Gemma 4 12B\citep{gemmateam2026gemma4technicalreport} & 12B & \texttt{google/gemma-4-12B-it} & native thinking \\
Gemma 4 26B-A4B & 26B (4B active) & \texttt{google/gemma-4-26B-A4B-it} & native thinking \\
Gemma 4 31B & 31B & \texttt{google/gemma-4-31B-it} & native thinking \\
\midrule
Claude Haiku 5.5 & undisclosed & Anthropic API, \texttt{claude-haiku-5-5} & adaptive, effort high, summarized \\
\bottomrule
\end{tabular}
\caption{Models. Open-weight models are served with vLLM at temperature 1.0 and top-$p$ 0.95; Claude Haiku uses API defaults. With reasoning off, thinking is disabled (K2 closes its thinking block at once) and answers are limited to 16 tokens. With reasoning on, the output budget is 12,288 tokens. Qwen and Gemma have no effort setting.}
\label{tab:models}
\end{table*}

\section{Compute}
All the experiments are done with 8 NVIDIA H200s. The experiments take around 2-3 GPU hours on average.

\section{Baseline Judgements by Model}
\label{app:acceptability}
\begin{figure}[h]
\centering
\includegraphics[width=\linewidth]{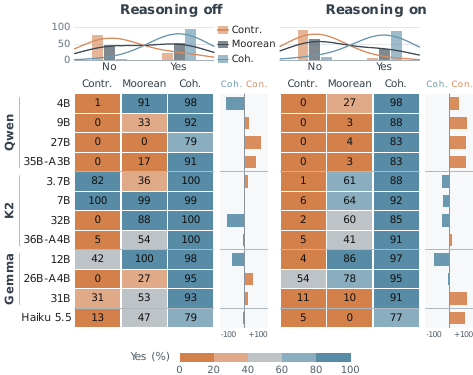}
\caption{Sensibility judgements per model, laid out as Figure~\ref{fig:baseline}. Side bars: $|M-\text{coherent}| - |M-\text{contradiction}|$, orange where the Moorean sentence is judged closer to the contradiction, blue where closer to the coherent sentence.}
\label{fig:baseline-models}
\end{figure}

\begin{figure}[h]
\centering
\includegraphics[width=\linewidth]{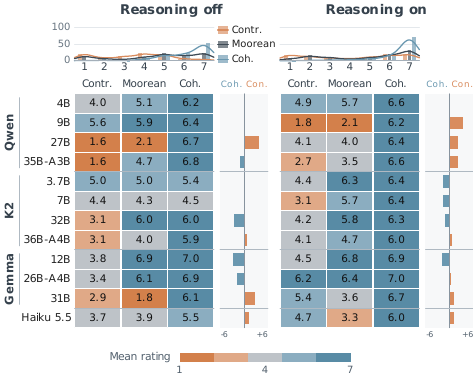}
\caption{Acceptability ratings (1-7) per model, laid out as Figure~\ref{fig:baseline-models}.}
\label{fig:baseline-rating}
\end{figure}

Figure~\ref{fig:baseline-models} gives each model's sense judgements of Section~\ref{sec:baseline}, and Figure~\ref{fig:baseline-rating} each model's acceptability ratings.

\section{Form Test by Model}
\label{app:form-models}
\begin{figure}[h]
\centering
\includegraphics[width=0.85\linewidth]{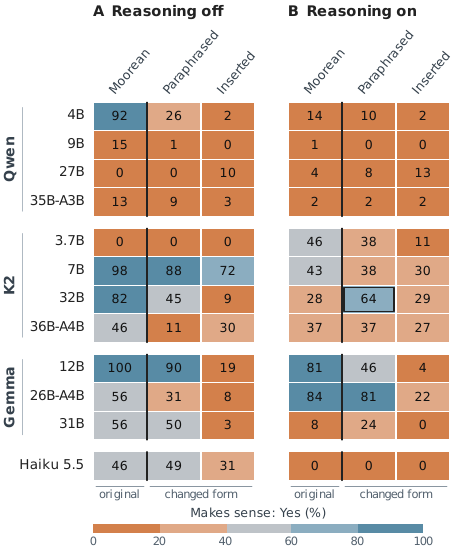}
\caption{Form test per model: share of Yes for the original Moorean sentence and its paraphrased (mean of three) and inserted (mean of two) versions, in the bands of Figure~\ref{fig:baseline}. Outlined: a changed form accepted although the original is rejected.}
\label{fig:form-models}
\end{figure}

Figure~\ref{fig:form-models} gives each model's results on the form test of Section~\ref{sec:form-test}.
Per model, the main exception is ``$p$, but I don't think it is'', which Gemma 31B and Haiku without reasoning accept more often than the original, possibly because the sentence reads as ``I think it isn't''. K2 32B with reasoning accepts every paraphrase more often than the original.

\section{Context Test by Model}
\label{app:context-models}
\begin{figure}[h]
\centering
\includegraphics[width=\linewidth]{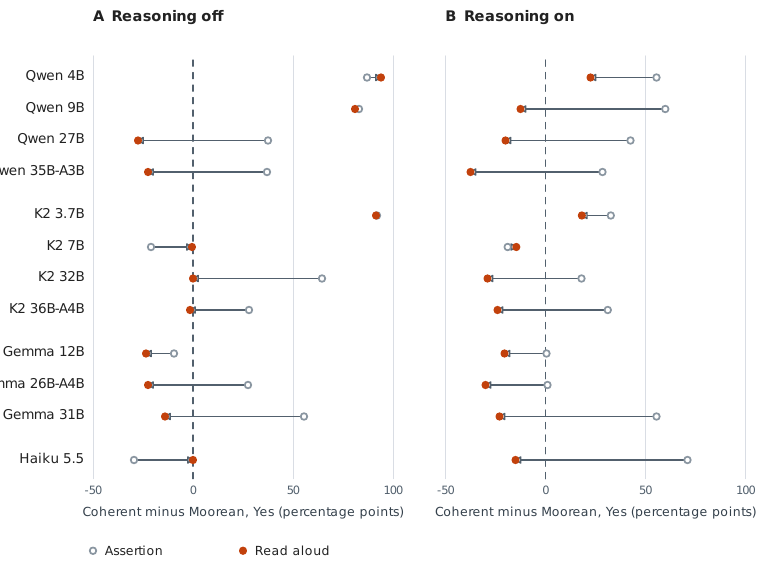}
\caption{Context test per model: the gap between the Yes rates for the coherent and the Moorean sentence when Robin asserts the sentence (open circle) and when she reads it aloud (filled circle). Below zero, the Moorean sentence is accepted more often than the coherent one.}
\label{fig:context-models}
\end{figure}

Figure~\ref{fig:context-models} gives each model's results on the context test of Section~\ref{sec:context-test}.

\section{Moore Mentions}
\label{app:moore}
\begin{figure}[h]
    \centering
    \includegraphics[width=0.75\linewidth]{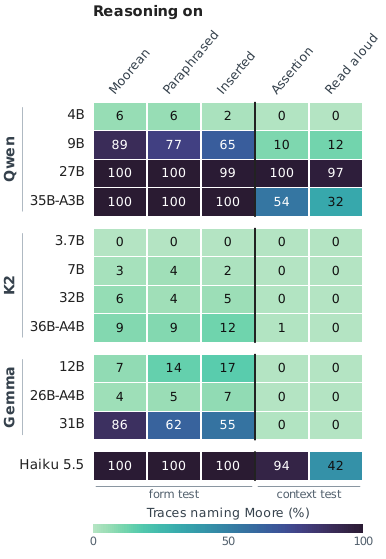}
    \caption{Share of reasoning traces that name Moore, for the Moorean sentence and its changed forms (form test) and for the Moorean sentence asserted or read aloud in the classroom setting (context test).}
    \label{fig:moore}
\end{figure}

Figure~\ref{fig:moore} gives the share of reasoning traces that name Moore, in the form test (Section~\ref{sec:form-test}) and the context test (Section~\ref{sec:context-test}).

\section{The Prior and the Reduction of $S_1$}
\label{app:prior}
\paragraph{$L_0$ with a prior.} Eq.~\ref{eq:l0} written out, with its normalizer over states, is
\[ \mbox{\scriptsize$\displaystyle L_0(s \mid u) = \frac{P(s)\,[\![u]\!](s)}{\sum_{s'} P(s')\,[\![u]\!](s')}.$} \]

\paragraph{Step 1. $P(s)$ cancels in $S_1$ under any prior.} Taking the log of $L_0$ gives
\[ \mbox{\scriptsize$\displaystyle \log L_0(s \mid u) = \log P(s) + \log [\![u]\!](s) - \log \sum_{s'} P(s')\,[\![u]\!](s').$} \]
$S_1$ normalizes $\exp(\log L_0(s \mid u)) = L_0(s \mid u)$ over utterances in a fixed state $s$,
\[ \mbox{\scriptsize$\begin{aligned}
S_1(u \mid s) &= \frac{L_0(s \mid u)}{\sum_{u'} L_0(s \mid u')} \\
&= \frac{\cancel{P(s)}\,[\![u]\!](s) \,/\, \sum_{s'} P(s')\,[\![u]\!](s')}{\cancel{P(s)}\,\sum_{u'}( [\![u']\!](s) \,/\, \sum_{s'} P(s')\,[\![u']\!](s'))}.
\end{aligned}$} \]
$P(s)$ is the same for every $u'$, so it comes out of the sum over $u'$ and cancels. The denominator is then the same for every $u$, and Eq.~\ref{eq:s1} becomes
\begin{equation} \mbox{\scriptsize$\displaystyle S_1(u \mid s) \propto \exp\Big(\log [\![u]\!](s) - \log \sum_{s'} P(s')\,[\![u]\!](s')\Big).$} \label{eq:s1-prior} \end{equation}

\paragraph{Step 2. The uniform prior turns the second term into a count.} With $P(s) = 1/|\mathcal{S}|$ for every state in the set $\mathcal{S}$ of states, $\sum_{s'} P(s')\,[\![u]\!](s') = \frac{1}{|\mathcal{S}|} \sum_{s'} [\![u]\!](s')$. Substituting into Eq.~\ref{eq:s1-prior} and normalizing over utterances,
\[ \mbox{\scriptsize$\begin{aligned}
S_1(u \mid s) &= \frac{[\![u]\!](s) \,/\, \frac{1}{|\mathcal{S}|}\sum_{s'} [\![u]\!](s')}{\sum_{u'} [\![u']\!](s) \,/\, \frac{1}{|\mathcal{S}|}\sum_{s'} [\![u']\!](s')} \\
&= \frac{\cancel{|\mathcal{S}|}\,[\![u]\!](s) \,/\, \sum_{s'} [\![u]\!](s')}{\cancel{|\mathcal{S}|}\,\sum_{u'} ([\![u']\!](s) \,/\, \sum_{s'} [\![u']\!](s'))}.
\end{aligned}$} \]
$|\mathcal{S}|$ is the same for every $u'$ and cancels in the same way, which gives the form of Section~\ref{sec:rsa},
\[ \mbox{\scriptsize$\displaystyle S_1(u \mid s) \propto \exp\Big(\log [\![u]\!](s) - \log \sum_{s'} [\![u]\!](s')\Big) = \frac{[\![u]\!](s)}{\sum_{s'} [\![u]\!](s')}.$} \]
$\sum_{s'} [\![u]\!](s')$ is the number of states where $u$ is true, so $1/\sum_{s'} [\![u]\!](s')$ is the informativity of $u$.

\paragraph{Eq.~\ref{eq:s1a} before and after the reduction.} The norms replace only the first term, $\log [\![u]\!](s)$, with $\log A(s,u)$. The informativity term still comes from $L_0$, so the prior enters as in Eq.~\ref{eq:s1-prior}. Under any prior,
\[ \mbox{\scriptsize$\displaystyle S_1(u \mid s) \propto \exp\Big(\log A(s,u) - \log \sum_{s'} P(s')\,[\![u]\!](s')\Big),$} \]
and under the uniform prior,
\[ \mbox{\scriptsize$\displaystyle S_1(u \mid s) \propto \exp\Big(\log A(s,u) - \log \sum_{s'} [\![u]\!](s')\Big) = \frac{A(s,u)}{\sum_{s'} [\![u]\!](s')}.$} \]
Where $A(s,u) = 0$, $\log A(s,u) = -\infty$, so $S_1(u \mid s) = 0$ and the norm rules $u$ out. Where $A(s,u) = [\![u]\!](s)$, Eq.~\ref{eq:s1a} is Eq.~\ref{eq:s1} under the same prior.

\paragraph{Example.} In $s_4$ with NA and JNA, $A(s_4,u)$ is 1 for $p$ and $M$ and 0 otherwise. $p$ is true in 3 states and $M$ in 2, so $S_1(p \mid s_4) = \frac{1/3}{1/3 + 1/2} = .40$ and $S_1(M \mid s_4) = \frac{1/2}{1/3 + 1/2} = .60$, as in Figure~\ref{fig:rsa}b.

\section{RSA with a Vanilla Speaker}
\label{app:vanilla}
\begin{figure}[h]
\centering
\includegraphics[width=\linewidth]{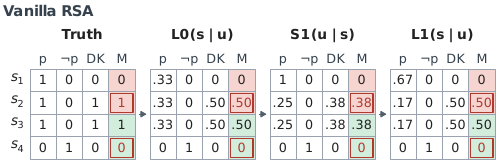}
\caption{RSA with a vanilla speaker, who may say anything literally true (uniform prior). Colours as in Figure~\ref{fig:rsa}.}
\label{fig:rsa-vanilla}
\end{figure}

Figure~\ref{fig:rsa-vanilla} gives the derivation of Section~\ref{sec:rsa} with a vanilla speaker.
A vanilla speaker $S_1$ considers, in each state (row), the utterances that are literally true there, and normalizes over them (Eq.~\ref{eq:s1}). In $s_2$, $p$, $\mathit{DK}$ and $M$ are true, so she may say any of them. The pragmatic listener $L_1$ considers, for each utterance (column), the states where $S_1$ would say it (Eq.~\ref{eq:l1}). This speaker misses the target of Table~\ref{tab:states} in two ways (Figure~\ref{fig:rsa-vanilla}). First, $M$ is literally true in $s_2$, where $p$ is true and Robin has no view, so she says $M$ there, although $M$ is infelicitous. Second, $M$ is literally false in $s_4$, where $p$ is false, so the reader never says $M$, although $M$ is felicitous there. The pragmatic listener inherits both errors. Hearing $M$, $L_1$ considers $s_2$, where $M$ should be ruled out, and assigns nothing to $s_4$, where $M$ should be possible.

\section{The Speaker Table $A(s,u)$}
\label{app:false-utt}
A norm of a speech act overrides cells of $A$ in two ways. It can forbid a true utterance, as JNA and BNA forbid $M$ in $s_2$, and it can permit a false one, as reading permits $p$ and $M$ in $s_4$ and BNA permits $\neg p$ in $s_3$. Since the speaker revises only the first term of her utility, a permitted false utterance keeps its informativity to the literal listener. $S_1(u \mid s) > 0$ exactly when $A(s,u) = 1$, which is the only property the tests of Section~\ref{sec:derivation} use.

For the 12 states of Section~\ref{sec:derivation}, $A_{\mathrm{Van}}(s,u) = [\![u]\!](s)$. $A_{\mathrm{NA}}$ equals $A_{\mathrm{Van}}$ except in the non-assertive states $s_4$, $s_{11}$ and $s_{12}$, where the norm of the act, not truth, decides what Robin may say. $A_{\mathrm{JNA}}$ and $A_{\mathrm{BNA}}$ equal $A_{\mathrm{NA}}$ except in the states where Robin asserts, where she may assert $u$ only if her evidence justifies it (JNA) or only if she believes it (BNA).

\section{Non-Assertive Acts}
\label{app:acts}
Non-assertive acts do not present the content as the speaker's own claim, and they differ in their norms. Quoting, acting a part or reciting resemble reading, while acts such as guessing or promising commit the speaker in other ways, and each act would bring its own norm and so its own instantiation of $A$. In Section~\ref{sec:rsa}, we instantiate the norm of non-assertive acts with reading, which requires saying what is printed and places no requirement on the speaker's epistemic state. The states of Section~\ref{sec:derivation} also include teasing ($s_{11}$) and misspeaking ($s_{12}$), which we treat in the same way, as acts without the epistemic requirement of assertion (the Non-assertive candidate). We define this group loosely for controllability; a finer-grained account would give each act its own norm.

\section{States}
\label{app:states}
Table~\ref{tab:model-states} describes the twelve states of Section~\ref{sec:derivation} by the world, Robin's evidence, belief and act.
\begin{table*}[tb]
\centering
\footnotesize
\setlength{\tabcolsep}{4pt}
\begin{tabular}{l p{7.2cm} llllcr}
\toprule
 & State & World & Evidence & Belief & Act & $M$ & Share (\%) \\
\midrule
$s_1$ & The world is $p$. Robin checked, and it looked $p$ (or a sign she trusts says $p$); she believes $p$. & $p$ & $p$ & $p$ & asserts &  & 20.0 \\
$s_2$ & The world is $p$. Robin has no evidence about $p$ and no view. & $p$ & none & none & asserts &  & 16.0 \\
$s_3$ & The world is $p$. Robin checked, and it looked $p$, but a feeling or emotion makes her believe $\neg p$. & $p$ & $p$ & $\neg p$ & asserts & \checkmark & 2.0 \\
$s_4$ & The world is $\neg p$. Robin's notes say $p$, and she is asked to read them aloud; she looked, and it did not look $p$; she believes $\neg p$. & $\neg p$ & $\neg p$ & $\neg p$ & reads ``$p$'' & \checkmark & 4.7 \\
\midrule
$s_5$ & The world is $\neg p$. Robin checked, and it did not look $p$; she believes $\neg p$. & $\neg p$ & $\neg p$ & $\neg p$ & asserts &  & 21.4 \\
$s_6$ & The world is $p$. From one side it looked $p$, from the other it did not; she has no view. & $p$ & both & none & asserts &  & 7.1 \\
$s_7$ & The world is $p$. Robin checked, and it looked $p$; she nonetheless believes $\neg p$. & $p$ & $p$ & $\neg p$ & asserts & \checkmark & 0.5 \\
$s_8$ & The world is $p$. A sign says $p$, but Robin does not trust it; she has not checked; she has no view. & $p$ & $p$ (sign) & none & asserts &  & 3.7 \\
$s_9$ & The world is $p$. A sign says $p$; her glance says $\neg p$; she finds $p$ more likely, yet believes $\neg p$. & $p$ & both & $\neg p$ & asserts & \checkmark & \multirow{2}{*}{6.6} \\
$s_{10}$ & As $s_9$, but she finds $\neg p$ more likely. & $p$ & both & $\neg p$ & asserts &  &  \\
$s_{11}$ & As $s_1$, and Robin teases a worried friend by pretending to doubt $p$. & $p$ & $p$ & $p$ & teases & \checkmark & 6.5 \\
$s_{12}$ & As $s_1$, but Robin, exhausted, meant to say another sentence. & $p$ & $p$ & $p$ & misfires & \checkmark & 0.4 \\
\midrule
 & Others (e.g., ambiguity in literal meaning, purpose only) & & & & & & 11.1 \\
\bottomrule
\end{tabular}
\caption{The states. Section~\ref{sec:rsa} uses $s_1$-$s_4$; Section~\ref{sec:derivation} uses all twelve, built from the models' proposals (Share: proportion of proposals mapped to each state).}
\label{tab:model-states}
\end{table*}

\begin{figure*}[ht]
\centering
\includegraphics[width=0.95\textwidth]{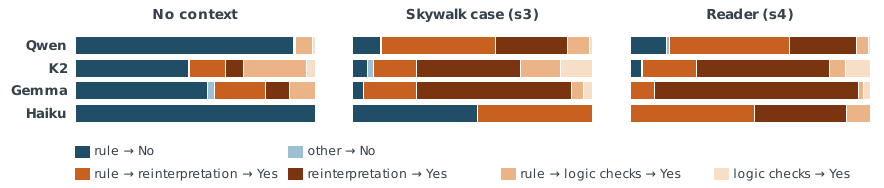}
\caption{Reasoning traces by the shape of their chain across three contexts. Per-model results in Appendix~\ref{app:traces}, Figure~\ref{fig:traces-models}.}
\label{fig:traces}
\end{figure*}


%

\section{Calibrating the Elicited Speaker/Listener}
\label{app:calibration}
We elicit $\hat{S}_1$ by asking the model to rate each utterance from 0 to 10 and normalizing the ratings over utterances. A model may rate an utterance Robin will never say 1 rather than 0, which leaves a small probability on that utterance. To determine which utterances $\hat{S}_1$ considers truly feasible, we calibrate $\hat{S}_1$ on two states where the ratings should be clear. In $s_1$, Robin's evidence shows $p$ and she believes $p$, so among the four alternatives she should say only $p$. In $s_5$, her evidence shows $\neg p$ and she believes $\neg p$, so among the alternatives she should say only $\neg p$. Each model's calibration baseline is the largest probability it gives to the wrong alternatives in $s_1$ or $s_5$ (the upper end of its 95\% bootstrap interval over propositions; .04-.15). We take $\hat{S}_1$ to consider an utterance feasible in a state if the lower end of the rating's interval exceeds the baseline, not feasible if the upper end is below it, and undetermined otherwise.

\section{Model Grids}
\label{app:model-grids}
Each model's $\hat{L}_0$, $\hat{S}_1$ and $\hat{L}_1$ on the 12 elicited states (reasoning on), with each $\hat{S}_1$ state-utterance pair classified against the model's hedging baseline (Section~\ref{sec:model-speaker}).

\begin{figure*}[p]
\centering
\includegraphics[width=0.85\textwidth]{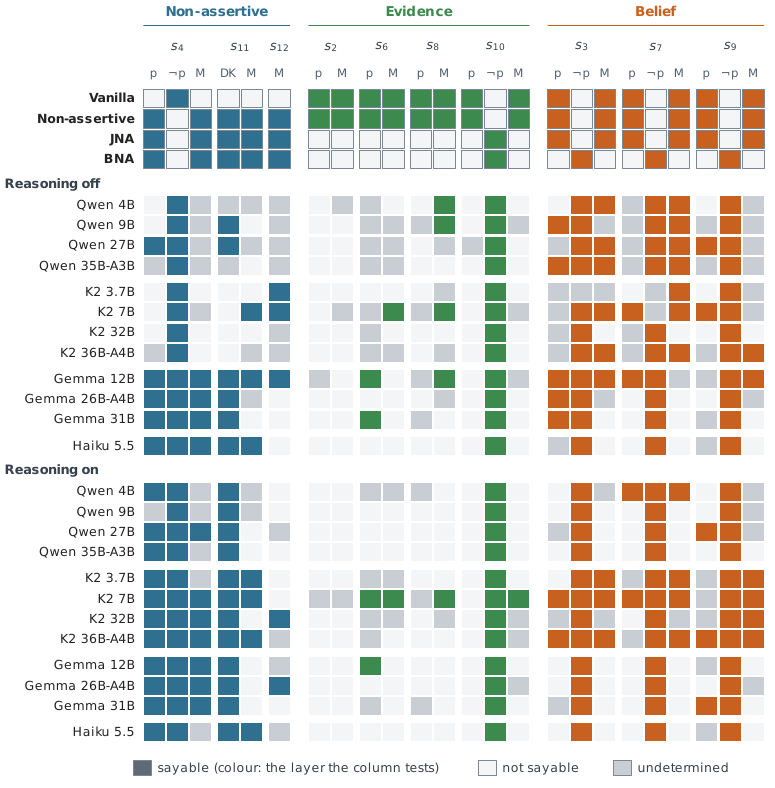}
\caption{The state-utterance pairs on which the candidate speakers differ, grouped by the layer that separates them, for the candidates (top) and each model without and with reasoning. Filled: feasible, in the colour of the layer; light: not feasible; grey: undetermined. Pairs on which all candidates agree are not shown. Agreement scores in Figure~\ref{fig:speaker-agreement}.}
\label{fig:speaker-tests}
\end{figure*}

\begin{figure}[tb]
\centering
\includegraphics[width=0.85\linewidth]{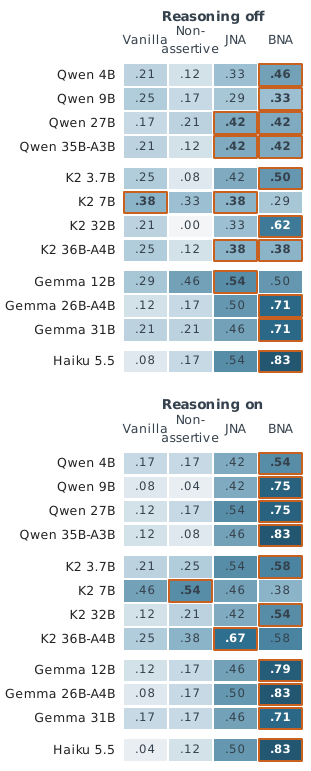}
\caption{Agreement of each model's speaker with each candidate (Eq.~\ref{eq:agree}), computed over the 24 diagnostic pairs on which the candidates differ (Figure~\ref{fig:speaker-tests}). On the other pairs, all candidates agree, so they shift every candidate's agreement equally. Undetermined pairs count as disagreements. Outlined: closest.}
\label{fig:speaker-agreement}
\end{figure}


\begin{figure*}[p]
\centering
\includegraphics[width=0.7\textwidth]{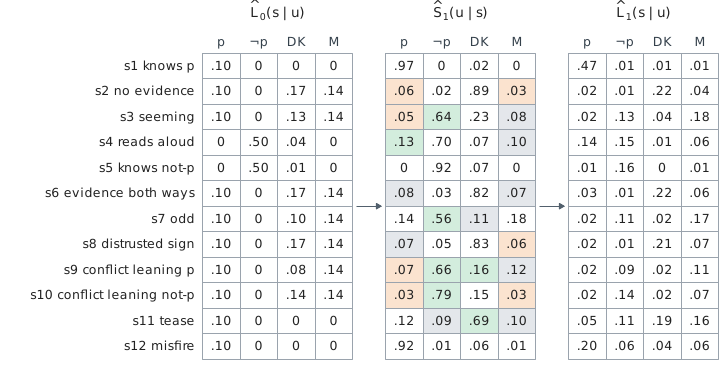}
\caption{Qwen 4B's grids. $\hat{S}_1$ is coloured against the vanilla speaker computed from the model's $\hat{L}_0$: green, feasible though the vanilla speaker would not say it; orange, not feasible though it would; grey, undetermined.}
\label{fig:example-grid}
\end{figure*}
\begin{figure*}[p]
\centering
\includegraphics[width=0.7\textwidth]{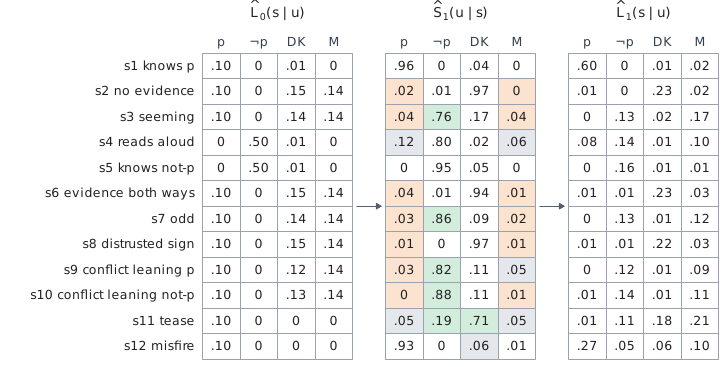}
\caption{Qwen 9B: grids, coloured as in Figure~\ref{fig:example-grid}.}
\end{figure*}
\begin{figure*}[p]
\centering
\includegraphics[width=0.7\textwidth]{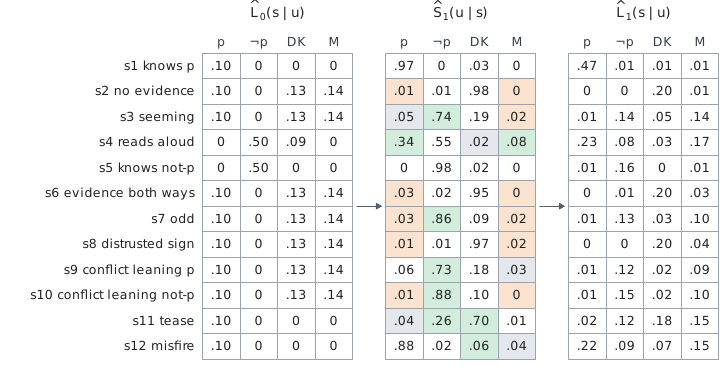}
\caption{Qwen 27B: grids, coloured as in Figure~\ref{fig:example-grid}.}
\end{figure*}
\begin{figure*}[p]
\centering
\includegraphics[width=0.7\textwidth]{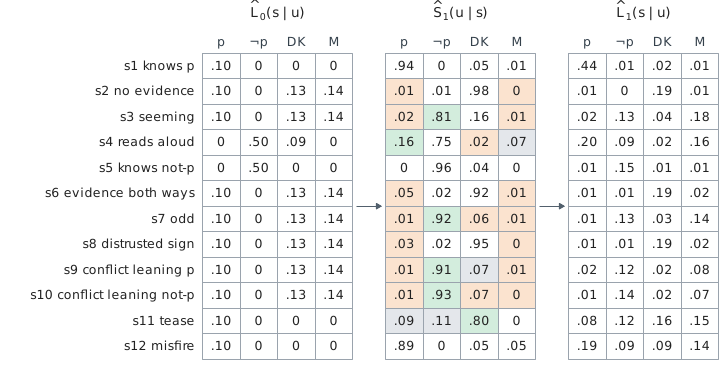}
\caption{Qwen 35B-A3B: grids, coloured as in Figure~\ref{fig:example-grid}.}
\end{figure*}
\begin{figure*}[p]
\centering
\includegraphics[width=0.7\textwidth]{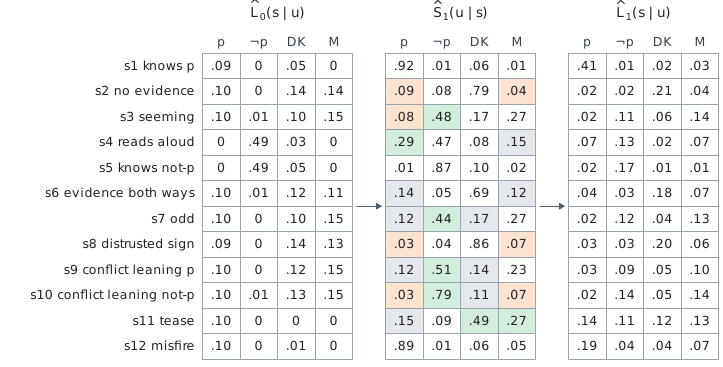}
\caption{K2 3.7B: grids, coloured as in Figure~\ref{fig:example-grid}.}
\end{figure*}
\begin{figure*}[p]
\centering
\includegraphics[width=0.7\textwidth]{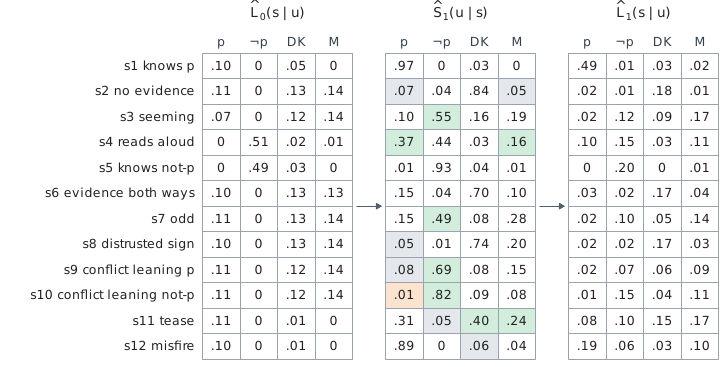}
\caption{K2 7B: grids, coloured as in Figure~\ref{fig:example-grid}.}
\end{figure*}
\begin{figure*}[p]
\centering
\includegraphics[width=0.7\textwidth]{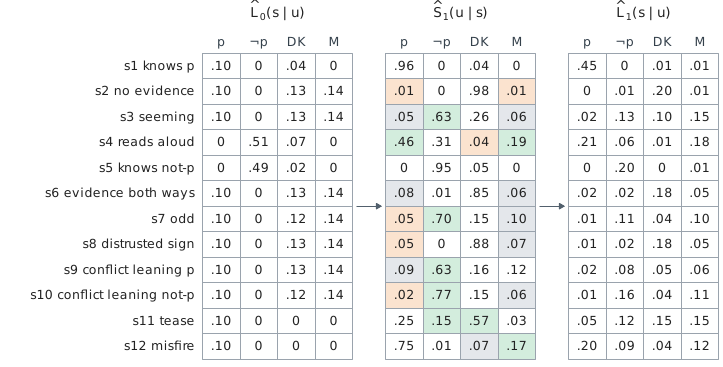}
\caption{K2 32B: grids, coloured as in Figure~\ref{fig:example-grid}.}
\end{figure*}
\begin{figure*}[p]
\centering
\includegraphics[width=0.7\textwidth]{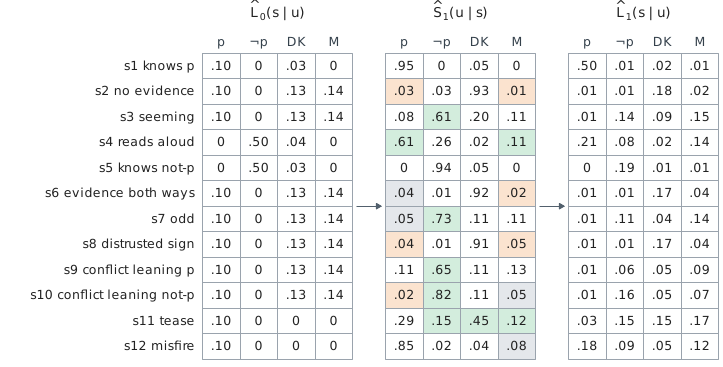}
\caption{K2 36B-A4B: grids, coloured as in Figure~\ref{fig:example-grid}.}
\end{figure*}
\begin{figure*}[p]
\centering
\includegraphics[width=0.7\textwidth]{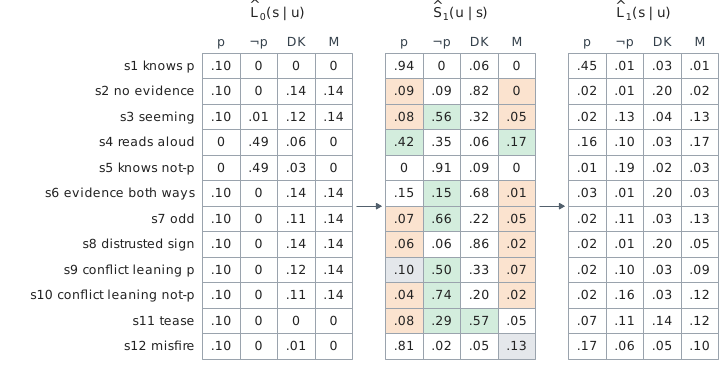}
\caption{Gemma 12B: grids, coloured as in Figure~\ref{fig:example-grid}.}
\end{figure*}
\begin{figure*}[p]
\centering
\includegraphics[width=0.7\textwidth]{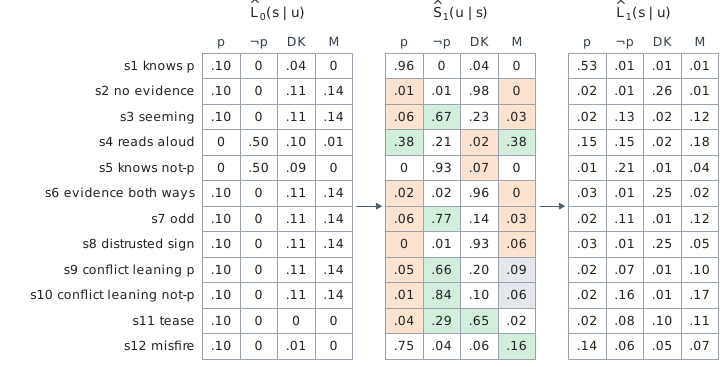}
\caption{Gemma 26B-A4B: grids, coloured as in Figure~\ref{fig:example-grid}.}
\end{figure*}
\begin{figure*}[p]
\centering
\includegraphics[width=0.7\textwidth]{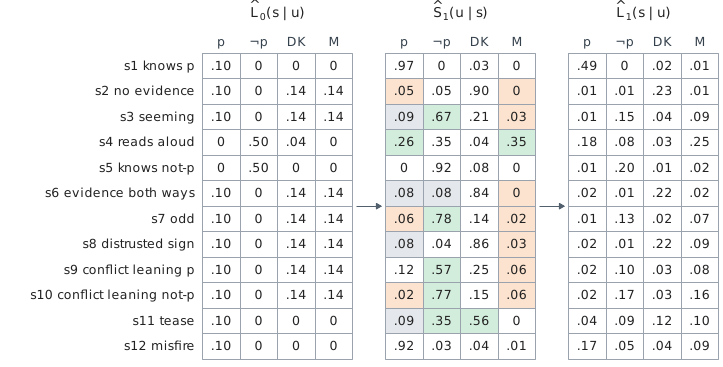}
\caption{Gemma 31B: grids, coloured as in Figure~\ref{fig:example-grid}.}
\end{figure*}

\begin{figure*}[p]
\centering
\includegraphics[width=0.7\textwidth]{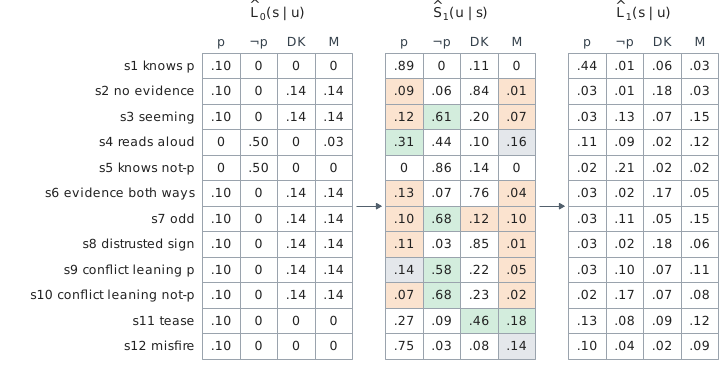}
\caption{Haiku 5.5: grids, coloured as in Figure~\ref{fig:example-grid}.}
\end{figure*}

\begin{figure}[p]
\centering
\includegraphics[width=0.85\linewidth]{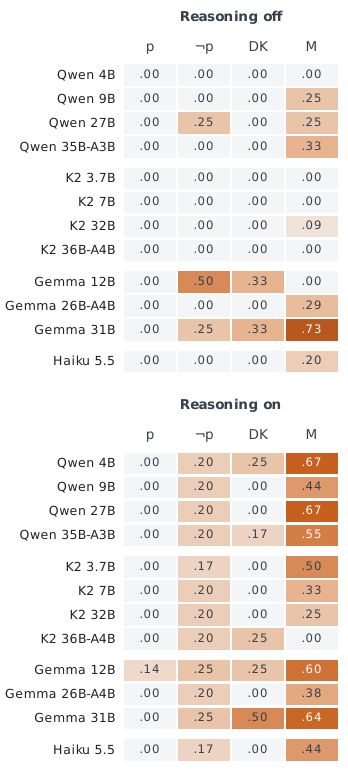}
\caption{False positives for every utterance: the share of the states where the speaker would not say it that the listener considers possible.}
\label{fig:listener-unfollowed}
\end{figure}

\begin{figure}[p]
\centering
\includegraphics[width=0.85\linewidth]{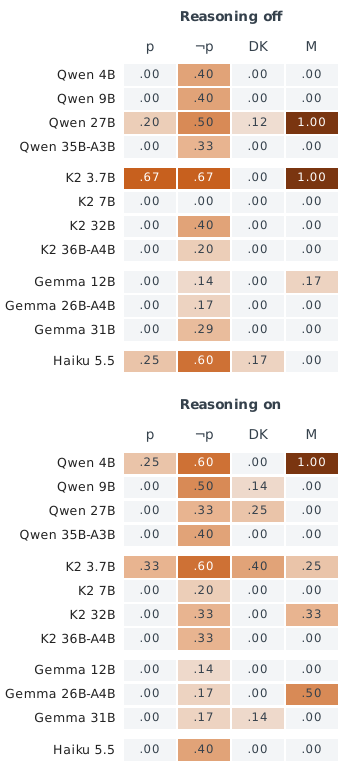}
\caption{False negatives in context for every utterance: the share of the contexts where the model's speaker would say it in which the model still judges it not to make sense.}
\label{fig:listener-fn}
\end{figure}


\section{Testing the Listener against Its Speaker}
\label{app:fn}
The listener weighs each state by the prior and by its speaker, $L_1(s \mid M) \propto P(s)\,S_1(M \mid s)$. A false positive therefore cannot come from the prior. Where the speaker rules out $M$, $S_1(M \mid s) = 0$, so $L_1(s \mid M) = 0$ under any prior, and a listener that still considers $s$ departs from its speaker. A false negative is different. Out of context, a low prior $P(s)$ alone can make the model reject $M$ even where its speaker would say $M$, so a rejection out of context does not show a departure. We therefore test false negatives in context. For each state where the model's speaker would say $M$, we give the model the state without its belief sentence and ask whether what Robin says makes sense (prompt in Appendix~\ref{app:prompt-grids}). For false negatives, we read $\hat{L}_1(s \mid M)$ from this accommodation in context, with $\hat{L}_1(s \mid M) = 0$ if the model, given $s$, accepts $M$ less than half the time. Undetermined states are left out of both rates. Figure~\ref{fig:listener-fn} gives the results for every utterance and context.

\newpage

\section{Into the Reasoning Traces}
\label{app:reasoning-traces}
Section~\ref{sec:derivation} shows that the models' judgements do not follow from reasoning over the speaker's state of the kind RSA describes. To further explore what the models rely on, we turn to their reasoning traces. The traces may also explain why the reasoning mode lowers the acceptance of Moorean sentences far more for Qwen and Haiku than for K2 and Gemma (Section~\ref{sec:baseline}). The traces give access to the models' explicit judgements \citep{zaman2025}, but prior work shows that they are not always faithful to model computation \citep{turpin2023, levy2025}. We therefore treat the traces as further behavior to analyze, rather than as a probe of the underlying mechanism \citep{zheng2026sda}.

\paragraph{Materials} We collect 1741 reasoning traces for the Moorean sentence from three settings, each over the same 50 propositions: out of context (from Section~\ref{sec:baseline}), the skywalk case ($s_3$), and the reader ($s_4$; both from Section~\ref{sec:model-listener}). 

\paragraph{Finding the reasoning chains.} We use an LLM annotator (DeepSeek V4.1 Flash; prompts in Appendix~\ref{app:prompt-traces}) in three steps. First, it summarizes each trace as an ordered list of two to six reasoning steps. Second, it groups these steps into types, which yields an inventory of 13 moves (Appendix~\ref{app:traces}, Table~\ref{tab:moves}). The inventory thus comes from the traces rather than from a taxonomy fixed in advance. Third, it labels each trace with the inventory to obtain the reasoning chains.

Exact chains rarely repeat, since they differ in length and in incidental moves. Several moves, however, play the same role in judging the sentence. Testing whether the sentence is a formal contradiction and separating the truth of $p$ from Robin's belief in it, for instance, both check whether the two halves of the sentence can hold together. We therefore group the moves manually by the role they play, yielding three major moves \emph{rule} (naming Moore's paradox or stating that sincere assertion implies belief), \emph{reinterpretation} (another reading of the sentence, such as an inner conflict or reading aloud) and \emph{logic checks} (testing whether the two halves can hold together). The chains then fall into six shapes (Figure~\ref{fig:traces}; grouping and examples in Appendix~\ref{app:traces}).

\paragraph{Results.} Figure~\ref{fig:traces} shows how the reasoning traces reach their judgements across the three settings. In the infelicitous setting (no context), the models reject the Moorean sentence mainly by invoking the rule that marks it as odd (the dark blue \emph{rule $\to$ No} bars, 70\% of chains). In the felicitous settings (the skywalk case and the reader), they accept it mainly by reinterpreting it, with or without invoking the rule first (the orange and brown bars), e.g., taking the first half as ``a report [of the notes], not an assertion of fact'' for the reader. The models thus rely on two things, a known rule about assertion or the paradox that marks the sentence as odd, and a reinterpretation that makes it acceptable. No move in any chain goes from Robin's evidence and belief to what she may assert, which is consistent with Section~\ref{sec:model-listener}.

Some models hold to the rule more stubbornly than others. In the skywalk case, Claude Haiku invokes the rule in every chain and lets it decide in about half (.52), rejecting the sentence although the context makes it acceptable. Without context, Qwen and Haiku likewise let the rule decide almost every chain (dark blue, .91 and 1.00), whereas K2 and Gemma often set it aside and reinterpret the sentence or check its logic (orange and beige). This is why, with reasoning, Qwen and Haiku reject the Moorean sentence while K2 and Gemma still accept it about half the time (Section~\ref{sec:baseline}).


\section{Reasoning Traces}
\label{app:traces}
\begin{figure*}[tb]
\centering
\includegraphics[width=0.95\textwidth]{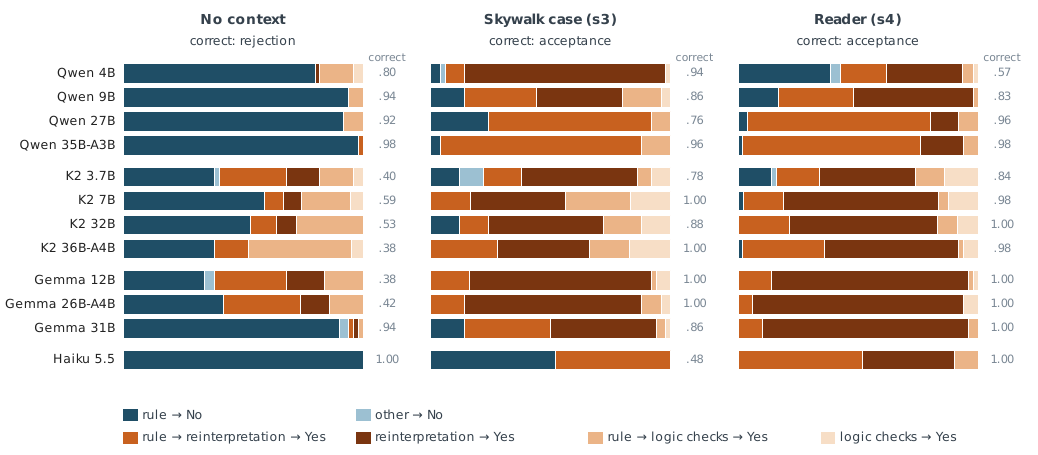}
\caption{Each model's reasoning traces by the shape of their chain, out of context, in the skywalk case ($s_3$), and for the reader ($s_4$). Blue: rejections; orange: acceptances. Right: share correct.}
\label{fig:traces-models}
\end{figure*}

Figure~\ref{fig:traces-models} gives each model's reasoning traces by the shape of their chain, Table~\ref{tab:moves} the 13 moves, and Table~\ref{tab:chains} the share of each chain shape in each setting.

\begin{table}[h]
\centering
\footnotesize
\setlength{\tabcolsep}{3pt}
\begin{tabular}{@{}p{0.80\linewidth}r@{}}
\toprule
Move & Traces \\
\midrule
States the verdict & .92 \\
Parses the sentence into a claim and a denial of belief & .73 \\
Restates the scenario (Robin's evidence, notes, feelings) & .59 \\
Checks for a logical contradiction & .57 \\
States that sincere assertion implies belief & .48 \\
Names Moore's paradox & .47 \\
Reads the sentence as expressing an inner conflict & .34 \\
Reads the first half as reading or quoting & .25 \\
Considers other readings (irony, doubt, reference of \emph{that}) & .22 \\
Separates truth or evidence from belief & .21 \\
Separates asserting from believing & .20 \\
Asks what \emph{makes sense} means & .16 \\
Distinguishes belief from alief or gut feeling & .03 \\
\bottomrule
\end{tabular}
\caption{The 13 moves found by grouping the reasoning steps, with the share of traces that make each.}
\label{tab:moves}
\end{table}

\begin{table}[h]
\centering
\footnotesize
\setlength{\tabcolsep}{4pt}
\begin{tabular}{@{}lccc@{}}
\toprule
Chain & No context & $s_3$ & $s_4$ \\
\midrule
rule $\to$ No & .70 & .11 & .06 \\
rule $\to$ reinterpretation $\to$ Yes & .11 & .32 & .31 \\
reinterpretation $\to$ Yes & .06 & .41 & .52 \\
rule $\to$ logic checks $\to$ Yes & .12 & .10 & .05 \\
logic checks $\to$ Yes & .01 & .06 & .04 \\
\bottomrule
\end{tabular}
\caption{The shapes of the reasoning chains, with the share of traces in each setting (other rejections: .01).}
\label{tab:chains}
\end{table}

\paragraph{Example chains.} One trace per chain shape, with the annotator's steps.

\noindent\textbf{rule $\to$ No} (Haiku 5.5, no context, No):
\begin{enumerate}\itemsep0pt
\item Identifies the statement as an instance of Moore's paradox.
\item Notes that such assertions are absurd despite not being logically contradictory.
\item Concludes that sincere utterance of the statement does not make sense.
\end{enumerate}

\noindent\textbf{rule $\to$ reinterpretation $\to$ Yes} (K2 36B-A4B, no context, Yes):
\begin{enumerate}\itemsep0pt
\item Parsed the statement into a factual assertion and a belief report, noting the 'but' suggests contrast.
\item Recognized that asserting a proposition typically implies belief, creating a tension with the second clause.
\item Considered that sincere utterances can honestly report or describe a fact while expressing personal disbelief, like 'It's raining, but I don't believe it.'
\item Concluded that the statement is coherent because it can express skepticism or surprise without logical contradiction.
\end{enumerate}

\noindent\textbf{reinterpretation $\to$ Yes} (Gemma 12B, $s_4$, Yes):
\begin{enumerate}\itemsep0pt
\item Identify the conflicting information: the notes say the service entrance is unlocked, but the actual state and Robin's observation show it is locked.
\item Interpret Robin's statement as having two parts: a report of what the notes say ('The service entrance is unlocked') and an expression of her own belief ('but I don't believe that').
\item Recognize that the context of reading notes aloud justifies the first part as a report, not an assertion of fact.
\item Conclude that the sentence coherently conveys both the source information and Robin's personal disbelief, making it sensible.
\end{enumerate}

\noindent\textbf{rule $\to$ logic checks $\to$ Yes} (K2 3.7B, no context, Yes):
\begin{enumerate}\itemsep0pt
\item The model identifies that the statement has two parts: a factual claim about the crate and a belief claim about that fact.
\item It questions whether there is a contradiction between the fact and the disbelief, noting that belief can be independent of objective truth.
\item It considers the role of sincerity and whether Robin knowing the fact would create inconsistency, but decides the context is unspecified.
\item It concludes that the statement is logically coherent because one can state a fact while not believing it.
\end{enumerate}

\noindent\textbf{logic checks $\to$ Yes} (K2 32B, $s_3$, Yes):
\begin{enumerate}\itemsep0pt
\item Interpreted Robin's statement as asserting the barrier is raised while denying belief in that assertion.
\item Considered whether asserting a proposition without believing it is coherent, noting that assertion does not always imply belief.
\item Examined the context of conflicting evidence: recent observation suggests raised, but past pattern suggests not raised.
\item Concluded that the statement is not self-contradictory and can make sense as an expression of uncertainty.
\end{enumerate}
\newpage
\clearpage
\section{Prompts}
\label{app:prompts}
Placeholders are in braces: \texttt{\{P\}} is the proposition as a sentence (e.g., ``The side gate is closed''), \texttt{\{p\}} the same in lower case, and \texttt{\{A\}}, \texttt{\{pred\}} another object of the item and the predicate. The \texttt{\{instruction\}} is ``Answer with exactly one word: Yes or No.'' without reasoning and ``Think it through, then end your reply with "Answer: " followed by Yes or No.'' with reasoning; in the baseline, the one-word instruction is used in both modes. With reasoning, the rating prompts end with ``Think it through, then end your reply with a JSON object, \dots''. The order of utterances and of states is shuffled.

\subsection{Propositions}
\label{app:prompt-propositions}
\codeboxinput{Proposition scenarios}{prompts/propositions.txt}

\subsection{Baseline}
\label{app:prompt-baseline}
\codeboxinput{Baseline prompt}{prompts/baseline.txt}
\codeboxinput{Baseline sentences and questions}{prompts/baseline_questions.txt}

\subsection{Form test}
\label{app:paraphrases}
The baseline prompt with the sense question ``Does what Robin says make sense?'', with one of the following sentences. \texttt{\{a\}}, \texttt{\{b\}}: two other objects of the item.
\codeboxinput{Form-test sentences}{prompts/form_sentences.txt}

\subsection{Context test}
\label{app:prompt-context}
\codeboxinput{Read aloud}{prompts/context_read.txt}
\codeboxinput{Assertion}{prompts/context_assertion.txt}

\subsection{Proposal Clustering}
\label{app:proposal_cluster}
We then use Claude Opus 5.5, which is not among the test models, to cluster each utterance's proposals. Then we manually map each group to a state with respect to the world, Robin's evidence, belief and act (Appendix~\ref{app:states}, Table~\ref{tab:model-states}). We set aside proposals that focus on lexical meaning that is less relevant to either speech act or speaker's epistemic state (e.g., ``that'' is anaphoric to something other than the proposition $q$, or the verb is ambiguous between two meanings, with a different meaning in each conjunct).

\subsection{State proposals}
\label{app:state-prompt}
\codeboxinput{State proposals}{prompts/state_proposals.txt}

\subsection{Model grids and accommodation}
\label{app:prompt-grids}
Each \texttt{\{state\}} is a scenario of Table~\ref{tab:model-states}.
\codeboxinput{Example state}{prompts/state_example.txt}
\codeboxinput{Literal meaning ($\hat{L}_0$)}{prompts/grid_literal.txt}
\codeboxinput{Speaker ($\hat{S}_1$)}{prompts/grid_speaker.txt}
\codeboxinput{Listener ($\hat{L}_1$)}{prompts/grid_listener.txt}
\codeboxinput{Accommodation}{prompts/accommodation.txt}

\subsection{Reasoning traces}
\label{app:prompt-traces}
\codeboxinput{Steps of a trace}{prompts/trace_route.txt}
\codeboxinput{Moves of a chain}{prompts/trace_chain.txt}

\end{document}